%% file: 0-main.tex
\documentclass[sigconf]{acmart}
\AtBeginDocument{%
  \providecommand\BibTeX{{%
    \normalfont B\kern-0.5em{\scshape i\kern-0.25em b}\kern-0.8em\TeX}}}

\setcopyright{acmcopyright}
\usepackage{multicol}

\usepackage{float}
\usepackage[utf8]{inputenc}
\usepackage{booktabs}
\usepackage{bbding}
\usepackage{pifont}
\usepackage{multirow}

\usepackage{amssymb}
\usepackage{natbib} 
\usepackage{graphicx}
\usepackage{pythonhighlight}
\usepackage{color}
\usepackage{appendix}
\usepackage{verbatim}
\usepackage{diagbox}
\usepackage[linesnumbered,ruled,vlined]{algorithm2e}
\usepackage{amsmath}
\usepackage{xspace}
\usepackage{subfigure}
\usepackage{caption}

\newcommand{\ignore}[1]{{}}

\newtheorem{definition}{Definition}

\usepackage{paralist}

\usepackage{accsupp}

\usepackage{listings}
\definecolor{keyword-green}{rgb}{0.2, 0.5, 0.3}
\definecolor{string-red}{rgb}{0.7, 0.1, 0.2}
\definecolor{important-red}{rgb}{0.8, 0.25, 0.2}
\definecolor{comment-grey}{rgb}{0.4, 0.5, 0.6}
\definecolor{item-blue}{rgb}{0.25, 0.3, 0.85}
\definecolor{text-blue}{rgb}{0.25, 0.2, 0.4}
\definecolor{number-black}{rgb}{0.1, 0.1, 0.15}
\definecolor{another-orange}{rgb}{0.7, 0.5, 0.15}

\begin{document}

% \title{A Fine-turning method for Time-series Foundation Models in Energy Forecasting Tasks via Curriculum Learning}
\title{Enabling Time-series Foundation Model for Building Energy Forecasting via Contrastive Curriculum Learning}

% Or
% Advancing Energy Forecasting with Time-series Foundation Models through Curriculum Learning

\author{Rui Liang$^*$, Yang Deng$^*$, Donghua Xie, Fang He, Dan Wang}

\affiliation{%
  \institution{The Hong Kong Polytechnic University 
%   $^2$Johnson Electric, 
%   $^3$University of California, San Diego
}
% %   \city{La Jolla}
% %   \state{CA}
}
\email{{maxwell-rui.liang@connect.polyu.hk, marco.deng@polyu.edu.hk}} 
\thanks{*Rui Liang and Yang Deng contributed equally to this research.}
% \email{{dan.wang,linda.xiao}@polyu.edu.hk}

% \renewcommand{\shortauthors}{Deng et al.}

\begin{abstract}

% \textbf{Style 1:}
% Conventional AI-based building energy forecasting schemes often suffer from their xxx (effort, large-scale, ). 
% Foundation models, i.e., the large deep learning neural networks can solve this problem because of its xx characteristics.
% However, .....

% \textbf{Style 2 (I think better):}
% \dy{
% 1 FM intro, 
% 2 general AI FMs there are many, 
% 3 There is a challenge to pushing general FM to building (lack of building knowledge). 
% 4 Then we need a Building FM to achieve xx
% }
% \dy{
% 5 In this paper, what's the goal, the challenge, and the solution.
% }

% \textbf{tell story:}
% 1. FM is import, emerging, and has been applied to various domains
% 2. Existing FMs perform not well on building. This is because building energy data is affect by many factors where each factor changes in different intervals. Therefore, building energy time-series exhibit unique multi-periodic patterns compared with other time-series and are hard to model.
% 3. To model such multi-periodicity, we propose ...

% % Large models, i.e., large language models (LLMs) or
% % Time-series foundation models (FMs) have transformed the way of cyber-physical systems. 

% % 需要 build 一个 BuildingFM的原因到底是什么：
% % IBM直接用去zero/few learning to a OIE 效果不行
% % 1 IBM don't learn many building knowledge because there are not too much source give it to learn
% % 2 

% (Long version)

Advances in time-series forecasting are driving a shift from conventional machine learning models to foundation models (FMs) that are trained with generalized knowledge.
However, existing FMs still perform poorly in the energy fields, such as building energy forecasting (BEF).
This paper studies the adaptation of FM to BEF tasks. We demonstrate the shortcomings of fine-tuning FM straightforwardly from both the perspectives of FM and the data.
To overcome these limitations, we propose a new \textit{contrastive curriculum learning}-based training method. Our method optimizes the ordering of training data in the context of TSFM adaptation. 
Experiments show that our method can improve the zero/few-shot performance by 14.6\% compared to the existing FMs.
Our code and new TSFM will be available at 
<Anonymous Github Repo>.
% https://github.com/anonymity/anonymity.

\end{abstract}

%   <concept>
%       <concept_id>10002951.10002952.10002971.10003450</concept_id>
%       <concept_desc>Information systems~Data access methods</concept_desc>
%       <concept_significance>500</concept_significance>
%       </concept>
%  </ccs2012>

% <concept_id>10011007.10011006.10011073</concept_id>
% <concept_desc>Software and its engineering~Software maintenance tools</concept_desc>
% <concept_significance>300</concept_significance>

% % \ccsdesc[500]{Information systems~Data access methods}
% \ccsdesc[300]{Software and its engineering~Software maintenance tools}

% <concept_id>10002944.10011123.10011130</concept_id>
% <concept_desc>General and reference~Evaluation</concept_desc>

% \ccsdesc[500]{General and reference~Evaluation}


% \settopmatter{printfolios=true}

\maketitle

\input{1-intro}

\input{2-related}

\input{3-solution}

\input{4-evaluation}

\input{7-conclusion}

% \input{8-acknowledgement}

% \newpage

\bibliographystyle{ACM-Reference-Format}

% \onecolumn
% \begin{multicols}{2}
\bibliography{ref}
% \end{multicols}

\newpage

% \input{9-appendix}
% \appendix

\end{document}

%% file: 1-intro.tex
\begin{figure}[t]
\setlength{\abovecaptionskip}{-1pt}
 \includegraphics[scale=0.28]{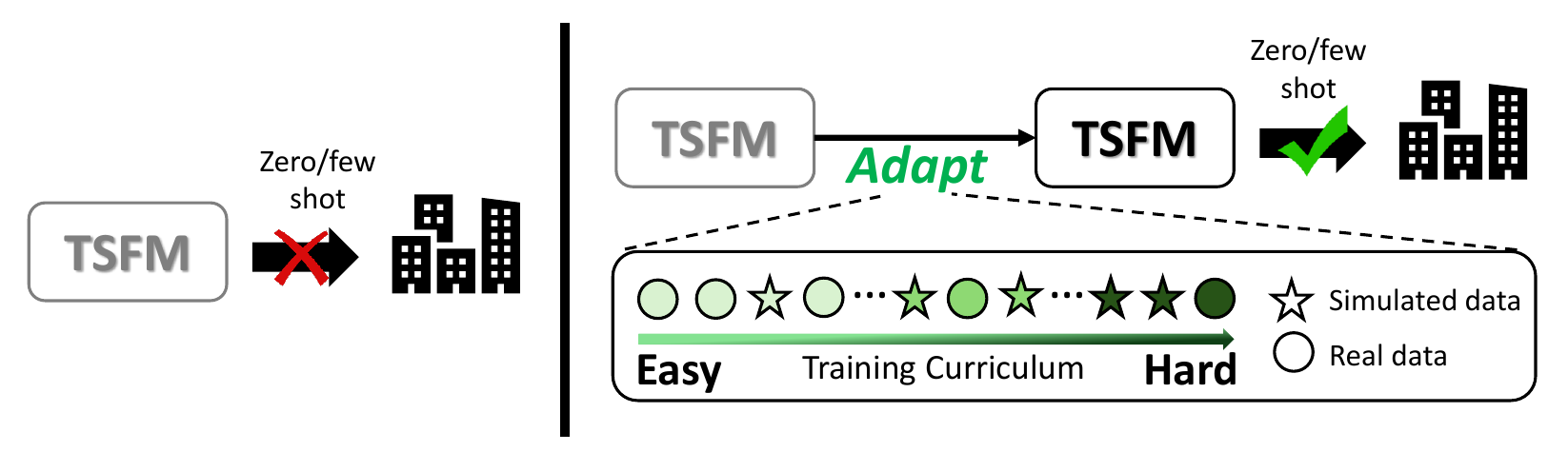}
\caption{ The curriculum enhances TSFM adaptation in building energy forecasting tasks.
% TSFM in building energy forecasting tasks. \textit{Left}: existing TSFM perform poorly; \textit{Right}: Our method.
}
\vspace{-3mm}
\label{fig: before-after}
\end{figure}

\section{Introduction}
\label{sec: intro}

Building energy forecasting (BEF), i.e., energy consumption forecasting for a building, plays a crucial role in many downstream applications, such as occupant behavior modeling.
% With the rapid progression of machine learning (ML) in building automation community, 
Currently, the majority of BEF schemes are based on machine learning techniques.
To achieve acceptable forecasting performance, the common practice is to develop specific models for individual buildings, yet it is hard to generalize at scale and requires significant effort.
A growing promising paradigm is
\textit{Foundation models} \cite{liang2024foundation}: large AI model trained on broad data such that it can be applied across a wide range of tasks, 
such as LLM in the NLP domain. 
% also known as the large pre-trained AI model, which is a deep learning model that has trained on broad data such that it can be applied across a wide range of use cases.  
Foundation models are capable of making inferences on a dataset with only a small fraction of training data, or even none at all, which corresponds to few-shot and zero-shot settings, respectively. 
% \todo{what is zero/few-shot}
% e.g., a prediction task, with or even without a little target data which corresponds to zero/few-shot learning.
% with or without minimal target data, i.e., corresponds to zero-shot and few-shot settings \cite{jin2023large}, respectively.
% \todo{what is zero-shot/few-shot}
Recently, various foundation models have been built in computer vision and language tasks to solve practical problems.
% such as automatic drive.
% , and the initial endeavors have been to develop the foundation models for time-series forecasting tasks (denoted as TSFM for simplicity). 

% The most limitation for this mode is the poor generalizability because such model is hard to perform well in other buildings since the data patterns are different among buildings. 
% % There are some transfer learning solutions \cite{} proposed to address this problem, yet many efforts needed to retrain the model and let along the performance.
% While there are transfer learning solutions \cite{li2021development} proposed to address this problem, they often require significant effort to retrain the model, not to mention the challenges related to performance.

% \todo{1. However, building energy data show unique pattern and are highly heterogeneous and existing TSFMs have not been trained on building energy data. Thus, TSFMs are not guaranteed to perform well in building energy. 2. Adaptation is a common method for enhancing the performance of TSFMs in a specific domain. 3. But public building energy data is limited.}
% \todo{1. TSFMs cannot perform well on building energy (not trained on building energy and building energy are heterogeneous) 2. Adapt is a common approach to enhance TSFMs on a specific domain 3. use simulated data bring new problem}

There are also some time-series foundation model (denoted as TSFM for simplicity) products available in 2024, such as IBM Granite \cite{ekambaram2024tiny} and Amazon Chronos \cite{ansari2024chronos}, with a series of TSFMs\footnote{TSFMs parameter size of IBM Granite are from 1M to 5M; for Amazon Chronos is range from 8.3M to 709M.} which are trained on various data source such as weather, energy, medical, financial.
From existing benchmarking \cite{liang2024foundation}, these TSFMs perform well in tasks, e.g., climate forecasting, where large-scale real datasets are adopted for pre-training.
However, there are limited real-world data resources in building scenarios and many BEF works rely on the simulated dataset. This is because the building energy is related to occupant privacy or business confidentiality and thus leads to the building managers having a low willingness to share the energy data \cite{xu2022understanding}.
From the recent measurement study \cite{mulayim2024time}, the existing TSFMs can not achieve acceptable performance in BEF.

% \dy{\textbf{Therefore}}, compared to other domains, 
% Thus, many BEF works rely on the \dy{simulated} dataset.
% (which with much larger size).

This paper is motivated by an essential question:
\textit{Can we adapt the existing TSFMs to support the building energy forecasting tasks via the currently available data resources?}
We perform a preliminary evaluation of day-ahead BEF on a product-level TSFM (under zero-shot setting). Our analysis compares the forecasting accuracy of 
i) the original pre-trained FM and 
ii) the FM fine-tuned using the BEF dataset in a straightforward manner. This fine-tuning can be conducted using either real-world data alone (R) or a combination of real-world and simulated data (R+S)\footnote{Real-world dataset: BDG \cite{miller2020building} consists of 1,000+ buildings; Simulated dataset: Building-900K \cite{emami2023buildingsbench} consists of 900,000 energy traces simulated by a business software EnergyPlus \cite{crawley2001energyplus}. All the buildings have a energy consumption time-series with a length of two-year.}. And half of the real buildings was allocated for the test set. 
As shown in Table \ref{tab: existing FMs performance},  the improvement by fine-tuning is very limited, and the performance can not reach engineering purposes. Specifically, even after incorporating 900,000 simulated buildings into the real dataset, the accuracy improved by only 0.6\%. 
% \dy{The results can yield two key insights: }
% (1) There are no so-called general patterns of building energy data, and patterns can \dy{vary greatly in difficulties} due to the diverse occupancy patterns and meteorological conditions.
% (2) \dy{Straightforward} training or fine-tuning can not enhance FM, even if the train set is large enough. This may be due to the knowledge learned in the pre-trained FM can not be quantified.
These results provide two key insights:
(1) There are no universal energy patterns for buildings, as patterns can vary significantly in complexity due to factors such as diverse occupancy behaviors and meteorological conditions.
(2) Straightforward training or fine-tuning does not enhance the FM, even with a sufficiently large training set. This is because the knowledge embedded in the pre-trained FM is not easily quantifiable.

% to adapt to BEF tasks.
% which \dy{especially works under the challenge of data scarcity} in the target domain. 

To address this challenging problem, we propose a new \textit{contrastive curriculum learning (CCL)} method to adapt the existing TSFMs to BEF tasks.
This curriculum strategy organizes the training samples in order of difficulty, thereby guiding the adaptation process of the TSFMs.
% We leverage the knowledge learned from the existing FMs instead of training an FM from scratch, which needs huge resources and efforts and does not suit the current stage. The contributions are as follows.
Instead of building an FM from scratch, which requires substantial resources and effort and is not feasible at this stage, we leverage the knowledge embedded in existing FMs. Our contributions are as follows:

\begin{itemize}
\item
We for the first time study the TSFM adaptation for a specific domain, building energy. And we demonstrate that straightforwardly fine-tuning brings limited gain.

% Our approach consists of three hierarchical curricula for assisting TSFMs in capturing the multi-periodic patterns of building energy data.

\item
% \todo{(produce the FM, we will open our FM after accept paper) We show that the propose fine-tuning approach can be adapted to TSFMs with different structures and sizes.}
We present a new contrastive curriculum learning method for adapting TSFMs to building energy forecasting tasks.
% , and the new TSFM will be public soon.
% and we will public new TSFM (will be publi)
% : \dy{(http://xxx)} which especially performs \dy{ well in education and lodging} buildings. 
% \dy{And we open-source: tinhou.}
% The checkpoints of the TSFMs will be open in Hugging Face.
% We \dy{build} TSFMs for typical BEC tasks, including short-term, medium-term, and long-term forecasting.
% daily, next point (or short-term / long term) and allows users to configure the xxx.

\item 
We evaluate the proposed method on three public building energy datasets.
Our evaluation indicates 9.9\% and 10.4\% overall zero-shot and few-shot performance improvement of our method as compared to direct fine-tuning.

\end{itemize}

\begin{table}[t]
    \caption{The BEF performance of TSFM from IBM. The metric is CVRMSE (lower the better) and 0.3 is acceptable for engineering purposes for BEF tasks.
    % $Acc$ is the performance of CV-RMSE (lower the better, and \dy{the acceptable threshold is 30\%}); $Acc_{+\mathcal{D}}$ denotes the performance of the FM after fine-tuned by dataset $\mathcal{D}$.
    }
    \vspace{-3mm}
    \centering
    \scalebox{.9}{
    \begin{tabular}{c|c|cc}
        % \hline 
        \toprule
        
       \textbf{Time-series FMs} & \textit{zero-shot} & \textit{FT on R} & \textit{FT on R+S} \\ \hline
        
        % Quality of Model &  & \checkmark & \checkmark & \checkmark  \\    

     IBM-Granite-TTM-5M  & 0.409 & 0.384  & 0.378 \\
        
    % Amazon-Chronos-T5-710M & 0.417 &   0.396  & 0.382 \\
        
         % Robustness Detection & \checkmark & \checkmark  &  &  \\
         % 之前这里是Adversarial Robustness
         
        \hline
        % \midrule

        % \textit{One customized LSTM} & \multicolumn{3}{c}{$Acc$: 24.5\%} \\

        \hline
        
    \end{tabular}
    }
    \label{tab: existing FMs performance}
    % \vspace{-2mm}
\end{table}

%% file: 2-related.tex
\section{Preliminaries}

\textit{Building Energy Forecasting.}
BEF is a domain-specific task of time-series forecasting: 
under the rolling forecasting setting with a fixed size window with a length of $L+T$, we have the data sample $u^t = (x^t,y^t)$ at time $t$, comprising past data $x^t = \{ l_1^t, ..., l_L^t \}$ with a look-back window length $L$ and future data $y^t = \{ l_{L+1}^t, ..., l_{L+T}^t \}$, where $l$ can be multi-dimension. 
Considering $y$ in BEF task is single dimension and currently published TSFMs mainly support univariate forecasting \cite{mulayim2024time}, thus $l \in \mathbb{R} $ in this paper.
We omit the superscript $t$ for simplicity later.

\textit{Curriculum Learning.}
Motivated by the feature of human education, curriculum learning is a data-centric training strategy in which an ML model is trained on samples of \textit{increasing difficulty} to smooth the training process and get better performance \cite{wang2021survey}. 
The two subtasks are: 
a \textit{Difficulty Measurer} to measure and rank the difficulty of samples; 
and a \textit{Training Scheduler} to decide the sequence of samples throughout the training process.

\textit{Contrastive Learning.}
Contrastive learning is to learn an embedding space to represent data samples, in which similar samples are grouped closer while dissimilar samples are pushed apart. 
The core is to construct positive pair $(u, u^+)$ and negative pair $(u, u^-)$ for an anchor sample $u$, i.e., to define the \textit{similarity}, based on which to train the NN-based encoder with a contrastive loss function.
% In conventional settings, negative examples are treated equally different from the anchors.
% To capture finer-grained information, Soft Contrastive Learning is proposed which explicitly differentiate negative samples by their similarity with the anchors.
% by contrasting positive and negative pairs of samples, i.e., $(\mathcal{S},\mathcal{S}^+)$ and $(\mathcal{S},\mathcal{S}^-)$. The core idea is to bring similar (positive) samples closer in the representation space while pushing dissimilar (negative) samples apart.

% \dy{Soft CL is xxx ----> seems say in 3.1 when use it.}

%% file: 3-solution.tex
\section{Methodology}

Problem statement:
given a pre-trained TSFM $\mathcal{M}$, 
the existing/available BEF datasets $\mathcal{D} = \{ {D}_{train}, D'_{train}\}$, where  ${D}_{train} = \{u\}$ is the real-world dataset, $D'_{train}=\{u'\}$ is the simulated dataset, and $\frac{|D_{train}|}{|D'_{train}|} \ll 1$.
Our objective is to adapt $\mathcal{M}$ to a new $\mathcal{M'}$ using $\mathcal{D}$, to minimize the loss of $\mathcal{M'}$ 
under zero/few-shot settings.

% a \dy{limited} training set $\mathcal{D}_{train}$ of building energy data, 
% a 
% our objective is to adapt $\mathcal{M}$ to a new $\mathcal{M'}$ using $\mathcal{D}_{train}$, to minimize the loss of $\mathcal{M'}$ on a testing building set $\mathcal{D}_{test}$ under zero/few-shot settings. 
% \dy{Note that the amount of data patterns involved in $\mathcal{D}_{train}$ is significantly smaller compared to the knowledge already acquired by $\mathcal{M}$.}

% \todo{Problem statement:}
% given a pre-trained TSFM $\mathcal{M}$, a training set $\mathcal{D}_{train}$ of building energy data, our objective is to adapt $\mathcal{M}$ to a new $\mathcal{M'}$ using $\mathcal{D}_{train}$, to minimize the loss of $\mathcal{M'}$ on a testing building set $\mathcal{D}_{test}$ under zero/few-shot settings. 
% \dy{Note that the amount of data patterns involved in $\mathcal{D}_{train}$ is significantly smaller compared to the knowledge already acquired by $\mathcal{M}$.}
% %  through a curriculum related to $\mathcal{D}_{train}$,

% Inspired by the progress of contrastive learning and curriculum learning, 
We propose a new Contrastive-aware Curriculum Learning (CCL) method to schedule the training process of $\mathcal{M}$ on $\mathcal{D}$, with samples ordered as \textit{easy-to-difficult} which is a common paradigm for ML model training, for example, \cite{wang2023efficienttrain} schedules the images from blur to clear to train the model.
A unique challenge in our scenario is to measure the difficulty of the simulated data. 
% which dominates the total $\mathcal{D}$ 
% but xxx without the reliable forecasting label $y$.}
We leverage contrastive representation to cope with this challenge, and the difficulty measurer and training scheduler designs are presented as follows.

% where the training set $\mathcal{D} = [(x_{i}, y_{i})]_{i}^{|\mathcal{D}|}$ contains both \dy{limited} real data and high-quality augmented data. 
% As a major contribution, the difficulty of augmented data can be well-conform via contrastive learning, thus our method is able to extend the training dataset in this task and let the TSFM be learned with a reasonable data order simultaneously. Fig x shows a diagram of the curriculum.
% The key designs of the proposed method are difficulty measurer $f_s$ and training scheduler $f_p$ are presented in the following sections.

% \dy{The overview of the proposed method is as follows (del):} 
% First, the difficulty of the real-world BEC data can be confirmed through the performance of the TSFM on the real data.
% we apply a traditional time-series augmentation approach to generate synthetic data by pattern-mixing, then a contrastive learning-based classifier is trained by the constructed sample pairs to judge the TSFM's mastery of the knowledge, based on which the di

% =====

\begin{figure}[t]
\setlength{\abovecaptionskip}{-1pt}
\includegraphics[scale=0.48]{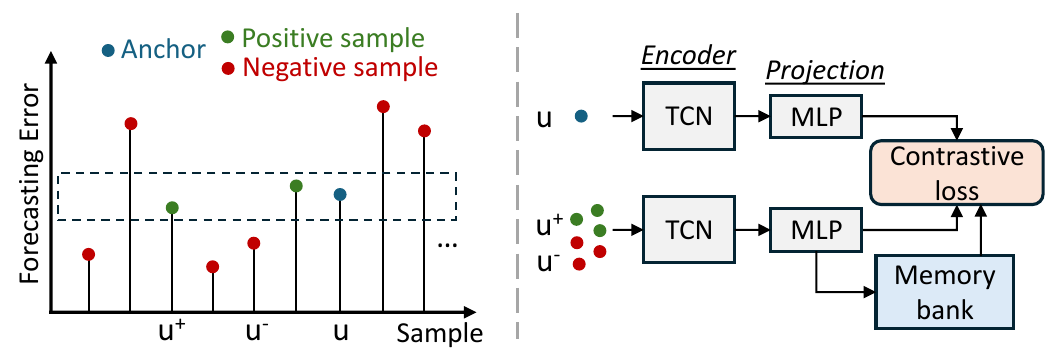}
\caption{Contrastive Learning model. \textit{Left}: contrastive pairs construction. \textit{Right}: NN model design.}
\vspace{-3mm}
\label{fig: CCL}
\end{figure}

\subsection{Contrastive-aware Difficulty Measurer}

% The difficulty measurer $f_s$ is different for the real-world and generated samples, denoted as $\{(x,y)\}$ and $\{(x', y')\}$ respectively.
% For real-world samples, 
The design of the difficulty measurer is usually based on model performance or data pattern analysis. Considering that the curriculum is for adapting an existing TSFM $\mathcal{M}$, which has been pre-trained with various knowledge and patterns,
we can directly make inference with the TSFM and use the performance as the \textit{difficulty score} of samples in the real-world dataset ${D}_{train}$ (Eq. \ref{definition: difficulty for real}).
Here, $\mathcal{L}(\cdot,\cdot)$ denotes the prediction error of the TSFM in terms of CV-RMSE.
\begin{equation}
    \label{definition: difficulty for real}
    D_{\mathcal{M}}(u)=\mathcal{L}(\mathcal{M}(x), y),
\end{equation}

% As mentioned, since the diversity and quantity of ${D}_{train}$ are limited to conduct TSFM adaptation, 
% Then, we study the suitable curriculum on $D'_{train}$.
For the simulated dataset $D'_{train}$,
% we \dy{first apply clustering-based pattern-mixing [must] (say earlier?)}, a conventional time-series augmentation method in energy scenarios, to generate \dy{high-fidelity (del)} synthetic data.
the key challenge is that the difficulty measurer for $u$ is not suitable for $u'$ because $\mathcal{L}(\mathcal{M}(x'), y')$ introduces bias. 
We then leverage contrastive learning to predict the TSFM comprehension on the representation of $u'$ and hence to determine the difficulty. 
We first define \textit{TSFM comprehension} and \textit{contrastive pairs},
% Then, we introduce the structure of the prediction model.
based on which we introduce the contrastive model and how to estimate the difficulty of $u'$.
% leverage contrastive learning to learn the \textit{TSFM comprehension} $r(u_1, u_2)$.
\begin{definition}
    \textit{TSFM comprehension on samples}. Let $C_{\mathcal{M}}(u_1, u_2) = |D_{\mathcal{M}}(u_1) - D_{\mathcal{M}}(u_2)| \in \mathbb{R}$ be the comprehension of a pre-trained TSFM $\mathcal{M}$ on two real sample $u_1$ and $u_2$. Less value denotes a similar comprehension of $\mathcal{M}$ on $u_1$ and $u_2$, and vice versa.
\end{definition}
\begin{definition}
    \textit{Contrastive pairs}. We define positive and negative contrastive pairs for real $u$ based on the value of $C_{\mathcal{M}}(\cdot, \cdot)$.
    Given an anchor $u$, the pair $(u, u^+)$ is positive if $C_{\mathcal{M}}(u, u^+) < \delta$ since $\mathcal{M}$ shows similar comprehension on the two samples; otherwise, the pair is negative, as $(u, u^-)$.
\end{definition}
Note that, contrastive pairs construction relies solely on real $\{u\}$ and we set $\delta$ to 0.01 based on our experimental analysis. 
It is a typical phenomenon that two dissimilar samples, which appear dissimilar in terms of time-series patterns (e.g., as measured by DTW similarity), may correspond to a similar TSFM comprehension, indicated by a low value of $C_{\mathcal{M}}(\cdot, \cdot)$. This is related to the uncertain knowledge encapsulated by the TSFM. 
% \rui{Based on our experimental analysis, we set $\delta$ to 0.01.}

The right part of Figure \ref{fig: CCL} shows the design of our contrastive learning model $f$.
Considering the huge amount of negative pairs, we leverage the classical memory bank structure and adopt temporal convolutional network (TCN) as the encoder since it can be trained efficiently and shows superior performance in capturing daily and weekly seasonality, which are major temporal patterns in building energy time-series. 
As shown in Eq. \ref{definition: contrastive loss}, 
we apply InfoNCE loss \cite{oord2018representation} and introduce the value of $C_{\mathcal{M}}(u, u_k)$ as the weight $\omega_k$ of negative pairs.
Here, $u_j$, $u_k$ are positive and negative samples of $u$, $sim(\cdot,\cdot)$ calculates the cosine similarity between each pair of data embeddings, and $\tau$ is a scaling parameter.
After training $f$, we obtain the difficulty of a simulated sample $u'$ through Eq. \ref{definition: infer the diffi}.
\begin{equation}
    \label{definition: contrastive loss}
    % \mathcal{L}_{cl} = \underset{u, u^+, u^-}{\mathbb{E}}\left\{-\text{log}\frac{\text{exp}(f(u)^{\top}f(u^+)/\tau)}{C_{\mathcal{M}}(u, u^-) \cdot \text{exp}(f(u)^{\top}f(u^-)/\tau)}\right\}
    \mathcal{L}_{cl} = -\text{log}\frac{\sum_{j=1}^{J}\text{exp}(sim(f(u), f(u_{j}))/\tau)}{\sum_{k=1}^{K}\omega_{k} \cdot \text{exp}(sim(f(u), f(u_{k}))/\tau)}
\end{equation}

\begin{equation}
    \label{definition: infer the diffi}
    % D(u') = D(\arg \min_{u \in \dy{U}}(\mathcal{L}_{cl}(u, u')))
    D_{\mathcal{M}}(u') = D_{\mathcal{M}}(\arg \min_{u \in D_{train}}(sim(f(u), f(u'))))
\end{equation}

\subsection{Training Scheduler}

% \dy{not mention "baby step" word (or other scheduler's name)? --- like: The widely used scheduler algorithm is baby step, which is xxx, then xx. To match our multi-xx curriculum, we design/calibrate xxx, xx. The overall is in Algorithm 1}
After measuring the difficulty of samples, we leverage a linear continuous scheduler to select training samples at each epoch.
Specifically, samples are first sorted by their difficulty.
% Then, a function $\lambda(t)$ is used to map the number of training epochs $t$ to a scalar $\lambda \in (0, 1]$ which represents that $\lambda$ proportion of the easiest samples should be used for training at the $t_{th}$ epoch.
Then, a function $\lambda(t)=\text{min}(1, \lambda_0+(1-\lambda_0)\cdot t/T_{\text{grow}} )$ decides the percentage of the easiest samples to be used at the $t$-th epoch, where $\lambda_0$ denotes the initial percentage of the easiest samples for training and $T_{\text{grow}}$ is the epoch when $\lambda(t)$ grows to 1.
Then, let $\mathcal{D}=\{v_i\}_{i=1}^{n}$, the training set at the $t$-th epoch is given by $\mathcal{D}_t=\{v_i\}_{i=1}^{n \cdot \lambda(t)}$.
In order to let TSFMs fully explore patterns in building energy data, we set the number of training epochs larger than $T_{\text{grow}}$, which means that TSFMs will be trained using the whole training set.

%% file: 4-evaluation.tex
\section{Evaluation}

\begin{table*}[ht]
\caption{Performance comparison (CV-RMSE, lower is better) under zero-shot and few-shot forecasting settings.
Improvement ratio of fine-tuned TSFMs as compared to original pre-trained TSFMs is shown at the last row.
% \rui{the number need match the number in table 1}
% with diverse horizons. The best results are highlighted in bold. 
% \todo{Can add an extra row for improvement ratio.}
}
\vspace{-4mm}
% \small
\centering
\scalebox{.77}{
\begin{tabular}{c|c|ccc|ccc|ccc|ccc}
\cline{1-14}
\multirow{3}{*}{Dataset} & \multirow{3}{*}{\begin{tabular}[c]{@{}c@{}} Forecast \\ horizon  \end{tabular}} & \multicolumn{6}{c|}{\textbf{Zero-shot setting}} & \multicolumn{6}{c}{\textbf{Few-shot setting}} \\ \cline{3-14} 
& & \multicolumn{3}{c|}{TSFM: TTM-5M} & \multicolumn{3}{c|}{TSFM: Chronos-710M} & \multicolumn{3}{c|}{TSFM: TTM-5M} & \multicolumn{3}{c}{TSFM: Chronos-710M} \\ \cline{3-14} 
& & Original & +FT & +\textbf{CCL-FT} & Original & +FT & +\textbf{CCL-FT} & Original & +FT & +\textbf{CCL-FT} & Original & +FT & +\textbf{CCL-FT} \\ 

% \midrule
\cline{1-14}
% \hline
\hline

\multirow{3}{*}{\begin{tabular}[c]{@{}c@{}} BDG-Fox \end{tabular}}
& 24 & 0.2175 & 0.2234 & \textbf{0.1952} & 0.1852 & 0.1735 & \textbf{0.1636} & 0.2183 & 0.2101 & \textbf{0.1947} & 0.1717 & \textbf{0.1566} & 0.1588 \\
& 96 &0.2942 & 0.2551 & \textbf{0.2367} & 0.2201 & 0.2174 & \textbf{0.1918} & 0.2616 & 0.2536 & \textbf{0.2244} & 0.2131 & 0.2147 & \textbf{0.1841} \\
& 192 &0.3306 & 0.2725 & \textbf{0.2565} & 0.2570 & 0.2446 & \textbf{0.2291} & 0.2773 & 0.2673 & \textbf{0.2582} & 0.2554 & 0.2477 & \textbf{0.2224} \\
\hline

\multirow{3}{*}{\begin{tabular}[c]{@{}c@{}} BDG-Rat \end{tabular}}
& 24 & 0.3094 & 0.3023 & \textbf{0.2714} & 0.2095 & 0.2078 & \textbf{0.1936} & 0.3097 & 0.2942 & \textbf{0.2628} & 0.2018 & 0.2024 & \textbf{0.1843} \\
& 96 & 0.4348 & 0.3687 & \textbf{0.3388} & 0.2468 & 0.2393 & \textbf{0.2033} & 0.3919 & 0.3624 & \textbf{0.3164} & 0.2294 & 0.2346 & \textbf{0.1996} \\
& 192 & 0.4582 & 0.3949 & \textbf{0.3676} & 0.2772 & 0.2516 & \textbf{0.2486} & 0.4204 & 0.4007 & \textbf{0.3555} & 0.2549 & 0.2324 & \textbf{0.2283} \\
\hline

\multirow{3}{*}{\begin{tabular}[c]{@{}c@{}} BDG-Bear \end{tabular}}
& 24 & 0.2514 & 0.2446 & \textbf{0.2103} & 0.1635 & 0.1611 & \textbf{0.1525} & 0.2546 & 0.2382 & \textbf{0.2080} & 0.1572 & 0.1583 & \textbf{0.1445} \\
& 96 & 0.3197 & 0.3009 & \textbf{0.2586} & 0.1919 & \textbf{0.1843} & 0.1860 & 0.3157 & 0.2996 & \textbf{0.2528} & 0.1739 & 0.1791 & \textbf{0.1718} \\
& 192 & 0.3162 & 0.3291 & \textbf{0.2827} & 0.2010 & 0.2125 & \textbf{0.1847} & 0.3083 & 0.3198 & \textbf{0.2779} & 0.1918 & 0.1883 & \textbf{0.1709} \\
\hline

\multirow{3}{*}{\begin{tabular}[c]{@{}c@{}} BDG-Panther \end{tabular}}
& 24 & 0.2736 & 0.2705 & \textbf{0.2482} & 0.1781 & 0.1657 & \textbf{0.1554} & 0.2708 & 0.2683 & \textbf{0.2423} & 0.1586 & 0.1487 & \textbf{0.1420} \\
& 96 & 0.3163 & 0.3068 & \textbf{0.2876} & 0.2154 & 0.1981 & \textbf{0.1612} & 0.3041 & 0.2852 & \textbf{0.2655} & 0.2039 & 0.1920 & \textbf{0.1464} \\
& 192 & 0.3259 & 0.3175 & \textbf{0.2983} & 0.2227 & 0.2168 & \textbf{0.2037} & 0.3142 & 0.3146 & \textbf{0.2914} & 0.2056 & 0.1923 & \textbf{0.1848} \\
\hline

\multirow{3}{*}{\begin{tabular}[c]{@{}c@{}} UCI \end{tabular}}
& 24 & 0.3848 & 0.3322 & \textbf{0.2761} & 0.2262 & 0.2035 & \textbf{0.1713} & 0.2981 & 0.3173 & \textbf{0.2666} & 0.2215 & 0.1966 & \textbf{0.1683} \\
& 96 & 0.3831 & 0.3415 & \textbf{0.2794} & 0.2564 & 0.2358 & \textbf{0.2106} & 0.3620 & 0.3376 & \textbf{0.2748} & 0.2425 & 0.2119 & \textbf{0.1937} \\
& 192 & 0.4074 & 0.3684 & \textbf{0.2955} & 0.2759 & 0.2664 & \textbf{0.2488} & 0.3743 & 0.3506 & \textbf{0.2906} & 0.2587 & 0.2544 & \textbf{0.2402} \\
\midrule

\multicolumn{2}{c|}{\begin{tabular}[c]{@{}c@{}} \textbf{\textit{Improvement ratio}} \end{tabular}}
 % & 7.8\% $\uparrow$ & 7.8\% $\uparrow$ & - & 7.8\% $\uparrow$ & 7.8\% $\uparrow$ & - & 7.8\% $\uparrow$ & 7.8\% $\uparrow$ & - & 7.8\% $\uparrow$ & 7.8\% $\uparrow$ & -  \\
 & - & 7.8\% $\uparrow$ & \textbf{18.3\% $\uparrow$} & - & 4.4\% $\uparrow$ & \textbf{12.6\% $\uparrow$} & - & 3.4\% $\uparrow$ & \textbf{14.9\% $\uparrow$} & - & 4.1\% $\uparrow$ & \textbf{12.7\% $\uparrow$} \\

\hline

\end{tabular}
}
\label{table: zero-shot & few-shot results}
\end{table*}

\begin{table}[ht]
\caption{Performance comparison of TSFM+CCL-FT and SOTA forecasting models in BEF field.}
\vspace{-4mm}
% \small
\centering
\scalebox{.75}{
\begin{tabular}{c|ccc|cc}

\hline
% \multirow{2}{*}{Dataset} & \multicolumn{2}{c|}{\textit{Zero-shot}} & \multicolumn{3}{c}{\textit{Full-shot}} \\ \cline{2-6}
Dataset & LSTM & Autoformer & TFT & TTM+\textbf{CCL-FT} & Chronos+\textbf{CCL-FT} \\ 
\hline

\multirow{5}{*}{}
BDG-Fox & 0.2797 & 0.3321 & 0.2259 & 0.2257 & 0.1884 \\
BDG-Rat & 0.4132 & 0.3407 & 0.2176 & 0.3115 & 0.2040 \\
BDG-Bear & 0.4308 & 0.4116 & 0.2093 & 0.2462 & 0.1624 \\
BDG-Panther & 0.2231 & 0.2506 & 0.3416 & 0.2664 & 0.1577 \\
UCI & 0.3592 & 0.1942 & 0.2125 & 0.2773 & 0.2007 \\
\cline{1-6}

\end{tabular}
}
\label{table: compare with SOTA}
\end{table}

\subsection{Methodology}

\textbf{TSFMs.}
We adopt two product-level TSFMs, which are Tiny Time Mixer (TTM) \cite{ekambaram2024tiny} from IBM and Chronos \cite{ansari2024chronos} from Amazon, for adapting to BEF tasks\footnote{We use the variants with the largest number of parameters for both TSFMs, which are TTM-5M and Chronos-710M.}.

\textbf{Datasets.}
Simulated and real-world public building energy datasets are used for experiments:
% \todo{Introduce the Buildings-900K dataset.}
(1) Buildings-900K \cite{emami2023buildingsbench}. This dataset contains hourly energy consumption time-series from 900k simulated buildings over two years.
% \todo{Only use a part of Buildings-900K? The size of this dataset is larger than pre-training set of TTM?}
(2) Building Data Genome Project (BDG) \cite{miller2020building}. 
This project aggregates 19 real-world building energy datasets from different locations around the world (totaling 1,636 buildings), where hourly energy meter data over a two-year period are collected for each building.
% Each dataset can be represented as BDG-xx where xx is the site name.
(3) UCI Electricity \cite{electricityloaddiagrams20112014_321}. This dataset collects electricity consumption data from 370 houses for four years, sampled at 15-minute interval.

\textbf{Baselines \& Metrics.} 
To explore improvement achieved by our method, we compare the TSFMs fine-tuned with our method (denoted as TSFM+CCL-FT) against:
(1) the original pre-trained TSFMs (denoted as TSFM);
and (2) the TSFMs directly fine-tuned without our method (denoted as TSFM+FT).
Besides, we adopt three state-of-the-art time-series forecasting models adopted in BEF field for comparisons: 
LSTM \cite{chitalia2020robust} (a classical RNN architecture for handling sequential data), 
Autoformer \cite{jiang2022very} 
and Temporal Fusion Transformer (TFT) \cite{giacomazzi2023short} (two transformer-based models tailored for time-series forecasting).
% \todo{Brief introductions of these baselines?}
For performance evaluation, we consider a standard metric in BEF field: the Coefficient of Variation of the Root Mean Square Error (CV-RMSE): $\frac{\sqrt{\sum_{i=1}^{n}(y_{i} - \hat{y})^{2}/n}}  {\sum_{i=1}^{n} y_{i} /n }$.

\textbf{Setup.}
We select five datasets from BDG together with the UCI dataset as the evaluation set since they cover most building types and climate conditions.
In zero-shot setting, all data from the target building are used for testing.
In few-shot setting, 10\% of data are used for training and the remaining 90\% of data are used for testing.
% For each building in the evaluation set, the first 50\% of data are reserved for specific training and the second 50\% are used for testing.
% \todo{add a graph for this???}
The other 15 datasets from BDG and the simulated dataset Buildings-900K are used for TSFMs fine-tuning.
% Among previously introduced 20 real-world datasets, we select 5 of them in diverse climate zones that cover all building types (e.g., commercial) as the evaluation set and apply the remaining datasets together with the simulated dataset for TSFM fine-tuning.
% \todo{How much data from Buildings-900K are used for fine-tuning.}
% Among these datasets, we select 5 of them in diverse climate zones that cover all building types (e.g., commercial) as the evaluation set and consider the remaining 15 datasets as the fine-tuning set.
% 1. univariate, context/horizon length, 10% of data, hardware setting
% We fine-tune and evaluate TSFMs used in our experiments under univariate setting\footnote{As most public TSFMs are pre-trained under univariate or channel-independence settings.} using the APIs and hyperparameter settings from the open-source GitHub repository.
The number of fine-tuning steps is set to 1000. 
The look-back window length and forecast horizon is set by default values of TSFMs during fine-tuning.\footnote{512-96 for TTM and 512-64 for Chronos.}
For evaluation, we set three forecast horizons, i.e., 24, 96, 192, as TSFMs can adapt to different horizons.
% 2. hyperparameter settings
% There are two evaluation settings which are zero-shot and few-shot for TSFMs in our experiments.
% For each test building, we use the first 50\% of data as the training set and the remaining 50\% as the test set.
% In zero-shot setting, the model is directly tested on the test set without training.
% In few-shot setting, 10\% of data from the training set is used for model training.
% For evaluation, since both TSFMs support adaptive forecast length, we set the horizon to 24, 96, and 192.
% 4. hardware settings
The experiments are conducted on a Linux server with two NVIDIA GeForce RTX 4090 24GB GPUs.

\subsection{Performance Result}

\textbf{Overall performance.}
We evaluate our method and baselines in zero-shot and few-shot settings under three forecast horizons in Table \ref{table: zero-shot & few-shot results}.
% Overall results/improvements
Overall, TSFM+CCL-FT consistently outperforms TSFM and TSFM+FT in zero-shot setting, with average improvements in CV-RMSE at 18.3\%, 11.3\% for TTM and 12.6\%, 8.5\% for Chronos.
Similar results are observed in few-shot setting where our method surpasses baselines by 14.9\%, 11.9\% for TTM and 12.7\%, 8.9\% for Chronos.
% The number of settings where we help the original TSFMs achieve CV-RMSE < 30%
Besides, as CV-RMSE < 0.3 is an industrial requirement defined by ASHRAE \cite{ashrae2002ashrae} for deployable forecasting models, we observe that on each dataset, there are cases where our method successfully reduces the error of pre-trained TSFMs to less than 0.3.
% Improvement on TTM is greater than Chronos, this is because their pre-trained knowledge are different
Moreover, we find that although our CCL method enhances the performance of both TSFMs over their pre-trained version, the improvement ratio is higher on TTM.
This is likely due to that TTM has a smaller model size and there are less energy datasets in its pre-training set such that the effect of our method is significant.
% FT is slightly better than Original
Additionally, we notice that for Chronos, TSFM+FT only slightly outperforms TSFM by 4.3\% which is consistent with our preliminary experiment result on TTM.

Next, we compare TSFM+CCL-FT with three state-of-the-art baseline forecasting models.
Here, TSFM+CCL-FT is evaluated under few-shot setting while the baselines are first trained using the first 50\% of data from each test building and then tested on the remaining 50\% of data.
As shown in Table \ref{table: compare with SOTA}, we observe that Chronos fine-tuned with the CCL strategy outperforms the best of baselines on almost all datasets, with an improvement of 14.6\% on average.
For TTM, although its performance is improved with our method, it still lags behind the best baseline, particularly on the BDG-Rat and UCI datasets.
% To further test whether the TSFMs fine-tuned with our method achieve sufficiently high performance, we compare the few-shot results of \textit{TSFM+CLFT} to full-shot results of state-of-the-art baseline forecasting models in the BEC field.
% Here full-shot means that the three baseline models are specifically trained using all training data from each test building.
% As shown in Table \ref{table: compare with SOTA}, the average CV-RMSE of \textit{TTM+CLFT}, \textit{Chronos+CLFT}, and the best results of baselines are 0.2654, 0.1826, and 0.2140, respectively.
% Although TSFMs are only trained on 10\% of data from the test building, they achieve comparable or even better performance as the best results of baselines.

% We next evaluate the zero-shot performance of our CL strategy and baselines under three typical BEC forecasting application scenarios which are hourly, daily, and monthly based on the forecasting horizon \cite{zhang2021review}.
% Note that, as the sampling frequency of testing data is inconsistent, we set different forecasting horizons for them for this experiment.
% For example, for day-ahead forecasting, the horizon is set to 24 for \textit{Genome} and \textit{HKisland} and 96 for \textit{UCI} and \textit{Smart}.
% Table \ref{table: first horizon analysis} shows that
% our method outperforms the baselines by 17.26\%, 17.09\%, and 11.87\% on average under the three forecasting horizon settings.
% \todo{This suggests...}

% The performance results for the other two TSFMs: Chronos-46M and Chronos-200M are in Appendix \ref{sec:appendix}.

\begin{figure}[t]
\setlength{\abovecaptionskip}{-1pt}
\centering
\begin{minipage}{0.23\textwidth}
    \setlength{\abovecaptionskip}{-1pt}
    \centering
    \includegraphics[scale=0.17]{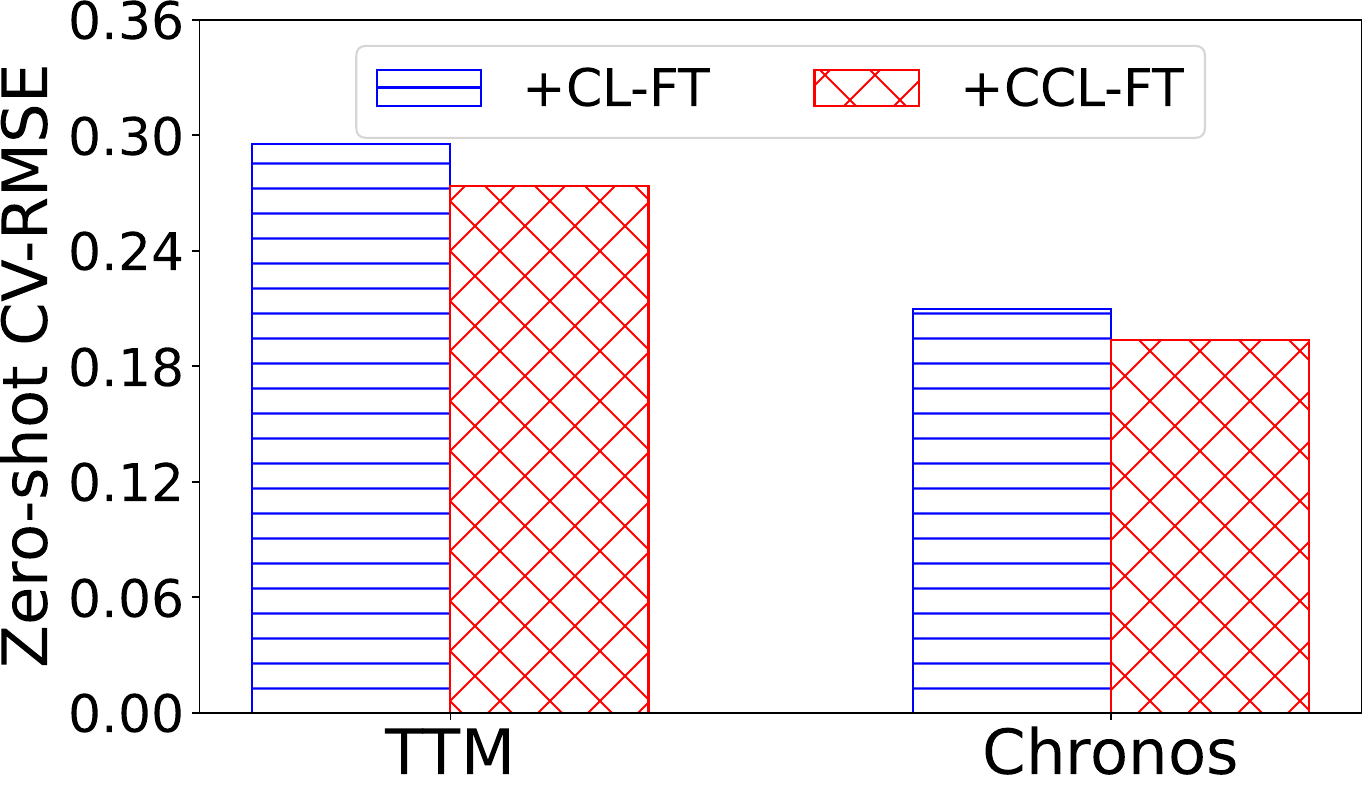}
    \caption{Comparison of CCL method and the variant.}
    \vspace{-2mm}	
    \label{fig: ablation}
\end{minipage}
% \hspace{5pt}
\hfill
\begin{minipage}{0.23\textwidth}
    \setlength{\abovecaptionskip}{-1pt}
    \centering
    \includegraphics[scale=0.17]{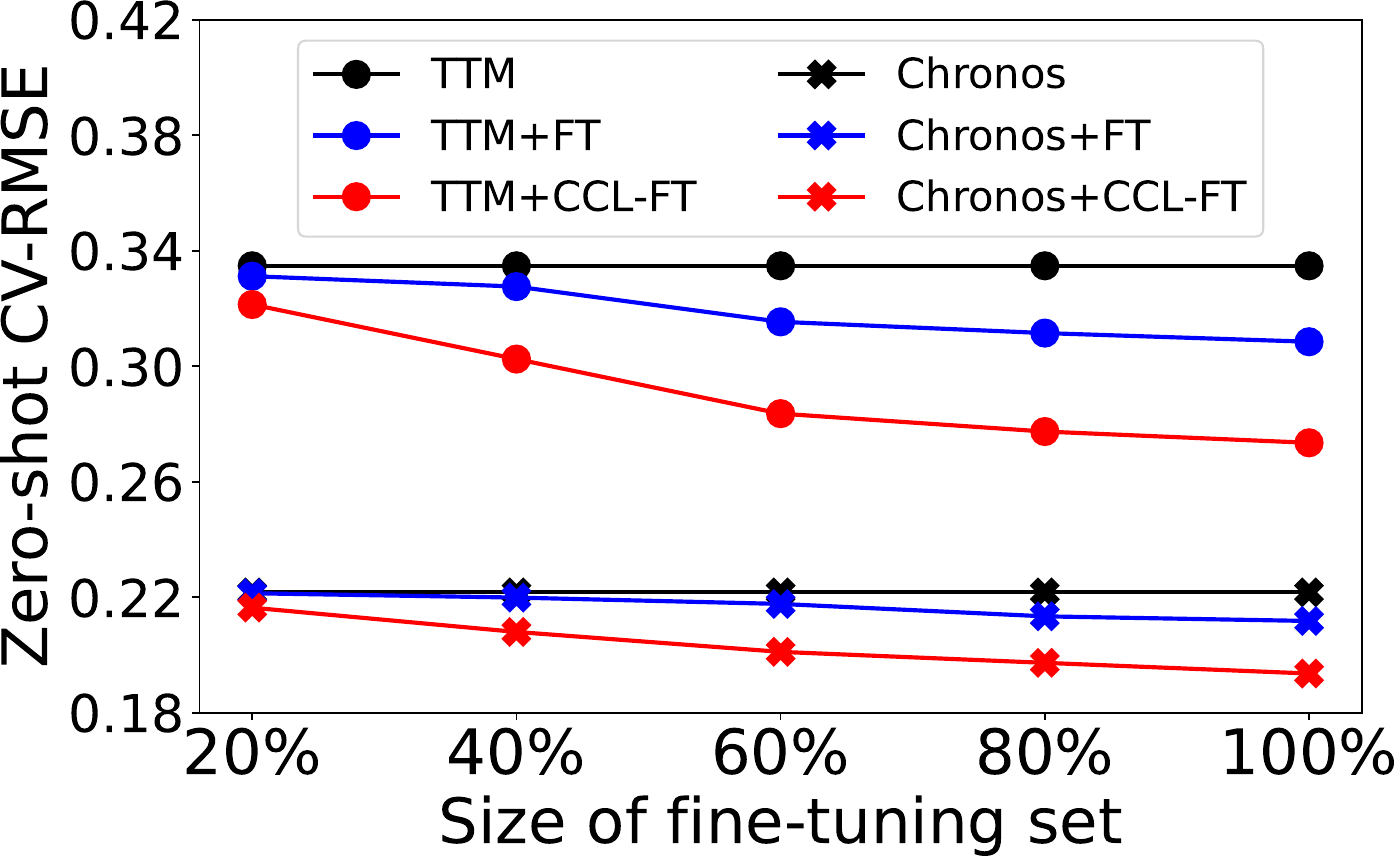}
    \caption{Comparison of different sizes of fine-tuning set.}
    \vspace{-2mm}	
    \label{fig: data size}
\end{minipage}
\end{figure}

% \subsection{Ablation Study}
\textbf{Ablation Study.}
To take a closer look at the contribution of the designed contrastive-aware difficulty measurer in our method, we implement a variant of our method named TSFM+CL-FT, which simply uses the performance of TSFMs as difficulty for both real and simulated samples.
The zero-shot performance of our method and this variant is compared in Figure \ref{fig: ablation}.
We observe that TSFM+CCL-FT outperforms TSFM+CL-FT by 7.4\% and 7.7\% for TTM and Chronos, respectively.
With further analysis, we find that TSFM+CL-FT is still better than TSFM-FT under the same experiment setting, which validates the effectiveness of curriculum learning in enhancing TSFMs adaptation.
% To take a closer look at the contributions of the three curricula in our method, we implement and examine three breakdown versions of the original CL method (denoted as $CL_{y,w,d}$) where each adopts only one of the three curricula (denoted as $CL_{y}$, $CL_{w}$, and $CL_{d}$).
% % We use $CL_{y,w,d}$ to represent the originally proposed method.
% As shown in Figure \ref{fig: ablation}, $CL_{y,w,d}$ consistently outperforms $CL_{y}$, $CL_{w}$, and $CL_{d}$ by 17.87\%, 16.01\%, and 11.27\%, respectively.
% This indicates the necessity of curricula for different periodicities.
% % patterns for the BEC forecasting task.
% Besides, we also observe 
% % that $CL_{d}$ achieves xx\% better performance as compared to $CL_{y}$ and $CL_{w}$.
% % This result may imply that 
% daily periodic patterns are more important for BEC tasks.

Next, we study the effect of the size of fine-tuning set on the performance of TSFMs.
In Figure \ref{fig: data size}, we evaluate the corresponding versions of fine-tuned TSFMs under varying proportion of fine-tuning set.
The performance of the original pre-trained TSFMs is included for reference.
The results indicate the superior performance of our method on each setting.
Specifically, we observe that the improvement over baseline increases along with the size of fine-tuning set from 2.9\% to 13.2\%.
This implies the capability of our method in handling a larger and more complicated dataset.

% We study the effect of the size of the training set on the performance of the three fine-tuning methods.
% As shown in Figure \ref{fig: data size}, we set up three training sets, i.e., \textit{Genome}, \textit{Genome}+\textit{UCI}, and \textit{Genome}+\textit{UCI}+\textit{Smart}.
% \textit{HKisland} is not included here due to its only nine buildings.
% The results show the CL method achieves superior performance under all three settings.
% Besides, we observe that the accuracy improvement between our method and the baselines increases as the size of the training set increases, which is from 4.62\% to 12.67\% and 15.06\%.
% This indicates that, with a larger and more diverse training set, our CL method exhibits stronger capability for TSFM fine-tuning.
% by providing a meaningful training order.

%% file: 7-conclusion.tex
\section{Conclusion}% and Future Work}

This work identifies a discrepancy between existing TSFMs and the in-use performance of TSFMs in a specific domain: building energy forecasting.
To bridge this gap, we have introduced a new curriculum learning-based training method in this context, to identify the difficulty of both real-world and simulated building data, and based on this to manage the order of train set during the process of TSFM adaptation.
% This study investigated the adaptation of existing time-series foundational models for building energy consumption (BEC) forecasting. 
% % Despite the effectiveness of recent FMs in various domains, their performance in the building energy sector has been limited due to scarce data and complex, multi-periodic consumption patterns.
% To address the challenge of scarce data resources, we fine-tuned existing FMs using curriculum learning, 
% % which optimizes the selection and sequencing of training data. 
The experiments show that the proposed curriculum design can greatly improve the zero/few-shot performance of two existing TSFM products among 1040 real buildings over three forecasting horizon settings. 
This work highlights the potential of curriculum learning to harness the generalized knowledge of FM tailored specifically for a specific domain.
% The future work will focus on performance improvement on certain contexts of buildings.
% % enhance BEC forecasting, offering more accurate and efficient energy management solutions.
% \todo{maybe future: the fidelity of the simulated dataset can guarantee? filter the low-quality ones? because this paper only rank the difficulty}

% \rui{where the definition of difficulty can be provided externally or discovered automatically. }